\documentclass{llncs}
\usepackage{makeidx}  
\usepackage{comment}
\usepackage{amsmath}
\usepackage{array}
\usepackage{graphicx}
\usepackage{longtable}
\usepackage{makecell}
\usepackage{cellspace}
\usepackage[binary-units = true,
            quotient-mode=fraction,
            group-minimum-digits=5]{siunitx} 
\DeclareSIUnit\px{px}
\DeclareSIUnit\fps{fps}
\usepackage{floatflt,epsfig} 
\begin{document}
\frontmatter          
\pagestyle{headings}  

\mainmatter              
\title{Efficient Semantic Segmentation for Visual Bird's-eye View Interpretation}
\author{Timo S\"amann \and Karl Amende \and
Stefan Milz \and \\ Christian Witt \and Martin Simon \and Johannes Petzold}
\authorrunning{Timo S\"amann et al.} 

\institute{Valeo Comfort and Driving Assistance, \\
Site Kronach (Germany), \\
Hummendorfer Str. 72,
96317 Kronach, \\
\email{timo.saemann@valeo.com}
}

\maketitle              

\begin{abstract}
The ability to perform semantic segmentation in real-time capable applications with limited hardware is of great importance. 
One such application is the interpretation of the visual bird's-eye view, which requires the semantic segmentation of the four omnidirectional camera images.
In this paper, we present an efficient semantic segmentation that sets new standards in terms of runtime and hardware requirements.
Our two main contributions are the decrease of the runtime by parallelizing the ArgMax layer and the reduction of hardware requirements by applying the channel pruning method to the ENet model. 
\end{abstract}
\section{Introduction}
\begin{figure}
\includegraphics[width=\textwidth]{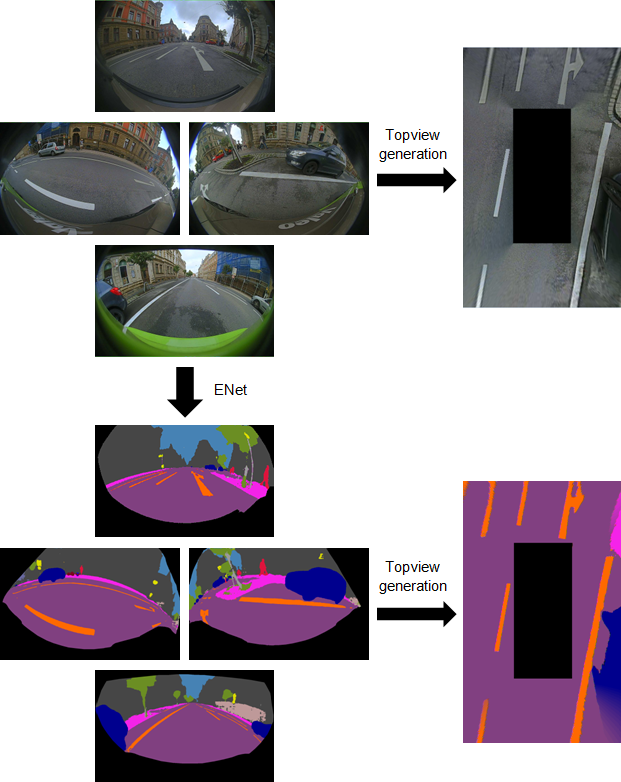}
\caption{Illustration of top view generation on raw fisheye camera and semantically segmented
images by projecting the images on a plane parallel to the ground using the camera model.}
\label{TopView}
\end{figure}

The understanding of scenes plays a key role in the technical realization of self-driving vehicles, home-automation devices and augmented reality wearables.
A prerequisite for understanding scenes based on cameras is the semantic segmentation. The aim of semantic segmentation is to classify every pixel of an image into meaningful classes. This task is typically realized with Deep Neural Networks (DNNs). The generation of a top view of a vehicle by using four omnidirectional cameras provides a $360^\circ$ surrounding bird's-eye view. The perception of the fully surrounding environment is important in many traffic situations for automated driving, e.g. autonomous parking. As shown in Figure~\ref{TopView} the interpretation and understanding of such a surround view could be done by DNN based semantic segmentation.

To enable the operation of DNNs on low power devices such as embedded systems in real-time, they need to be implemented efficiently.
\cite{paszke2016enet} represents a Deep Neural Network architecture (ENet) for real-time semantic segmentation. The ENet is listed on the Cityscapes benchmark as the fastest model, while provides a respectable quality that is sufficient for many application~\cite{cordts2016cityscapes}.
However, we were able to show that the computational and memory requirements are too high for generating the semantically segmented top view image on the NVIDIA Jetson TX2 board in real-time.

In terms of runtime, the ArgMax layer represents a bottleneck on models for semantic segmentation. It determines the indices of the maximum values along the depth axis for the output feature maps. In most publications, this calculation is excluded from the runtime measurement~\cite{paszke2016enet,long2015fully,badrinarayanan2017segnet}, since this calculation is very time-consuming. In the case of a real-time application, this calculation is relevant and cannot be ignored.
By parallelizing the ArgMax calculation on the GPU, the runtime of this layer can be drastically reduced on the NVIDIA TX2 board compared to common CPU implementations. 

Since we have to calculate the semantic segmentation four times to generate the top view image and the available resources on embedded systems are scarce, we reduced the number of parameters and thus the required GPU memory of the ENet by a variant of the channel pruning method~\cite{he2017channel}. The idea of this method is to prune channels\footnote{The term \textit{channels} is synonymous with \textit{feature maps}.} of convolutional layer by a LASSO regression based channel selection followed by a fine-tuning step for recover the weights. We extend the idea of channel pruning for image classification to the task of semantic segmentation and prune the ResNet based network ENet. For our experiments, we used the ENet implementation of Caffe which is publicly available on GitHub\footnote{https://github.com/TimoSaemann/ENet}.

\section{Releated Work}
\label{Releated_Work}
Efficiency is one of the key research areas for automated vehicles. Since the real-time capability is needed it has become a mandatory requirement for DNN applications. 
ENet is one of the most efficient Deep Neural Networks for semantic segmentation~\cite{paszke2016enet}. The ENet consists of an encoder-decoder structure. Unlike SegNet~\cite{badrinarayanan2017segnet}, which uses a symmetric encoder-decoder structure, the ENet uses a larger encoder and a smaller decoder, which reduces the computational effort. 
Additionally, ENet places great importance on early reduction of input information. Calculation operations on input images with a lower resolution are less complex and require less time. 
Furthermore, it uses asymmetric convolution presented in~\cite{szegedy2016rethinking}. An $n\times n$~filter is divided into an $n\times1$ and $1\times n$~filter. Both filters applied one after another which results in the same output as a $n\times n$~filter once applied, with the advantage of lower computational effort.

A massive amount of work on DNN acceleration has been done in the following three fields~\cite{he2017channel}: 1. Optimized implementation~\cite{hessam2016}, 2. Quantization~\cite{DBLP:journals/corr/RastegariORF16} and 3. Structured simplification~\cite{DBLP:journals/corr/JaderbergVZ14}. However, the choice of the right method strongly depends on the application task and the basic architecture of the used DNN.

An optimized implemented method speeds up convolutions by special convolution operations or approximations. Similar to this, quantization tries to approximate large floating point multiplications by less complex reduced floatings points or single bit operations~\cite{DBLP:journals/corr/RastegariORF16}. There are methods for sparse connection~\cite{DBLP:journals/corr/HanPTD15} or tensor factorization~\cite{DBLP:journals/corr/JaderbergVZ14}, which decompose weights into subsets.

A famous method to improve residual block based architectures in terms of efficiency and memory consumption is channel pruning~\cite{he2017channel}. The basic idea of this method is to reduce the number of channels that serve a convolutional layer as an input while maintaining the output of the layer. This means that only channels are removed, which have a minor impact on the output. Those channels can be found by performing a LASSO regression. Formally, suppose we apply the filter $W$ with $n\times c\times k_h\times k_w$ to a sample of an input $X$ with $N\times c\times k_h\times k_w$, the output $Y$ results with an output size $N\times n$. The letter $c$ represents the number of channels, $n$ the number of output feature maps, $N$ the number of input samples and $k_h$, $k_w$ are the filter size. The channel pruning method can be described as follows:

\begin{equation*}
\label{prune_reg}
\arg\min\limits_{\beta, W} \frac{1}{2N}\left\Vert Y- \sum_{i=1}^c \beta_i X_i W_i^T\right\Vert^2_F  \textmd{subject to} \left\Vert \beta \right\Vert_0 \leq c'
\end{equation*}

$\left\Vert \cdot \right\Vert_F$ designates the Frobenius norm. $X_i$ and $W_i$ designates the input data and filter of a channel $c$ with $i = 1,\ldots,c$, respectively. $\beta$ is a vector of size $c$ and takes either 0 or 1 for each element. If $\beta_i = 0$, the channel with the corresponding index $i$ gets removed from the feature map. $\Vert \beta \Vert_0$ is less than or equal to $c'$, which represents the maximum number of remaining channels.

Channel pruning could be separated into training based methods and inference-time based methods \cite{he2017channel}, whereas the latter is extremely challenging especially for very deep architectures, e.g.~residual networks like ENet. \cite{he2017channel} mentions a bottom-up technique, where first channel pruning is applied to a single convolutional layer. Afterwards, the method is stretched to the whole model. The results mentioned in \cite{he2017channel} are promising for residual blocks.

The top or birds-eye view generation is a common state of the art feature in almost every 360 degree surround view application for advanced driver assistance systems \cite{Zhang2014ASV}. The basic idea is a texture mapping of four omnidirectional cameras, which are mounted with different viewing directions into a top view plane (see Fig.~\ref{TopView}). The aim of this feature is a better environmental perception. Therefore, \cite{1801.00708} fundamentally studied the semantic segmentation of such a top-view. This is helpful for freespace and road marking detection within automated driving applications. An efficiency investigation for such an application is missing. The aim of this work is to provide an efficient solution to enable semantic top view interpretation for automated driving.
\section{Methods}
In this section, we first propose an efficient ArgMax implementation to accelerate the forward pass of the ENet. Then we describe how we prune the ENet architecture using the channel pruning method.

\subsection{ArgMax Implementation}
\begin{figure}[htbp]
\centering
\includegraphics[width=0.55\textwidth]{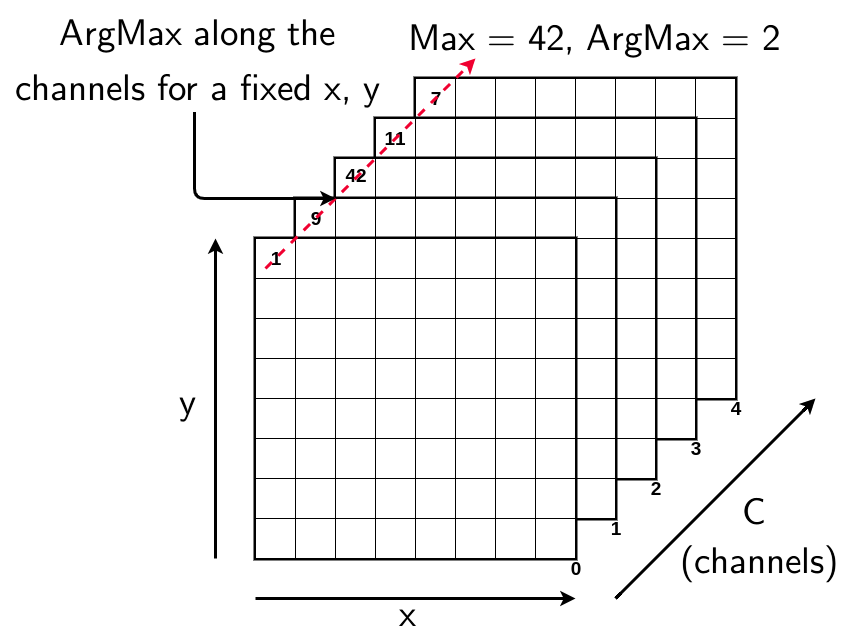}
\caption{Example of the ArgMax operation for one pixel.}
\label{Argmax:1}
\end{figure}

The ArgMax layer is the last layer in a model for semantic segmentation. For every pixel, the index of the maximum value along the depth axis is determined (see Fig.~\ref{Argmax:1}). The resulting index corresponds to the class that the pixel will be assigned to.
Due to the serial implementation on the CPU, which is used by Deep Learning Frameworks such as Caffe~\cite{jia2014caffe}, the layer becomes a bottleneck especially for embedded systems such as the NVIDIA TX2 board.
In order to get a high frame rate on embedded hardware, we need to implement the ArgMax layer on the GPU.

The ArgMax calculation for a pixel requires the values along the depth axis, as shown in Fig.~\ref{Argmax:1}. In theory, it is possible to calculate the ArgMax for all pixels simultaneously without conflicts.
We used this observation and implemented a GPU version for the ArgMax layer with CUDA in Caffe. We implemented our custom ArgMax kernel which calculates the ArgMax for a given pixel along the depth axis. Every CUDA thread computes the ArgMax for exactly one pixel. Since there are no dependencies or conflicts between the pixels during the ArgMax calculation, we can use the maximum number of threads and have to read each value from the input exactly once. In this way, we achieve a high degree of parallelism.
The results and speed of our implementation are presented in the section~\ref{sec:Results}, Table~\ref{tab:ArgMaxComparison}.

%
\subsection{ENet's Channel Pruning}

\begin{floatingfigure}[r]{3.7cm}
\mbox{\includegraphics[width=0.25\textwidth,height=90mm]{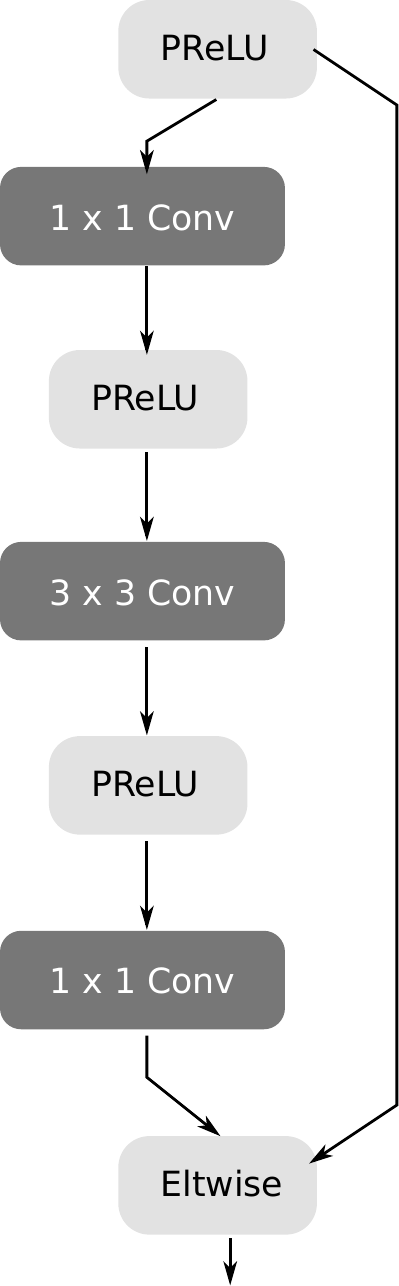}}
\caption{Example residual block as it occurs several times in the ENet architecture.}
\label{fig:residual_block}
\end{floatingfigure}

ENet's Channel Pruning consists mainly of two steps. The first step is to select the channels that can be pruned and the second step is the fine-tuning step where the weight parameters are recovered in a fine-tuning.

The ENet architecture is based on many consecutive residual blocks. An example residual block is shown in figure~\ref{fig:residual_block}. Please note that the batch normalization and dropout layer have been merged into the convolutional filters. In each residual block, every second and third convolutional (conv) layer was pruned. 

In the middle layer ($3\times3$~conv), the number of channels was reduced by a certain ratio. The depth of the filter of the following convolutional layer ($1\times1$~conv) has been adapted accordingly. The selection of the channels and filter depths to be pruned was done using the LASSO regression as described in section~\ref{Releated_Work}.

The first layer of the residual block was not pruned because for the practical implementation a so-called \textit{feature map sampling} layer has to be applied before the first convolutional layer.   

\cite{he2017channel} claims that the runtime of the \textit{feature map sampling} layer is negligible, but we found that the runtime is longer than the saved runtime due to the lower computational effort resulting from the smaller number of channels.

The number of channels to be pruned depends on the channel factor. This value is divided by the number of existing channels and thus determines the ratio of the pruned channels. The selection of the appropriate channel factor value is crucial for the success of the pruning. Therefore, the choice of this value is discussed in more detail in the next section~\ref{Experiments}. In the last residual block, no pruning was performed, because the number of channels is only 4 and a reduction would affect the quality very negatively.

In the fine-tuning step, we used again the customize training data set. In contrast to the previous training from the scratch, which passed through 150 epochs, the fine-tuning step can be limited to a few epochs (3 to 5) in order to achieve network convergence. Since the pruned ENet requires less GPU memory for training, the batch size can be increased from 6 to 11 per GPU. The learning rate was set to $10^{-8}$, which is ten times higher than the learning rate after 150 epochs of training from the scratch. The remaining training parameters have been adopted from~\cite{paszke2016enet}.

\section{Channel Factor Selection}
\label{Experiments}
In our experiments, we tested various channel factors which we use to reduce the feature maps.
We followed the approach of~\cite{he2017channel} and chose a high channel factor of 1.5 for the shallower residual blocks and a lower factor of 1.25 for the deeper residual blocks. 
We found out that the quality of the model decreased sharply and reversed the ratio of the channel factor. Now the quality of the model was only slightly reduced. This leads us to the assumption that the number of channels in the shallower residual blocks should not or only slightly be reduced. The reason for this might be, that the number of feature maps in the shallower residual blocks is lower than in the deeper ones. Therefore, we reduced the channel factor of the shallower residual blocks from 1.25 to our final value of 1.1. Since the number of channels in the shallower residual blocks is quite small (16), the benefit of a larger reduction is low. As a result, we find these values as a better compromise between saving computational effort and losing quality.

\section{Results}
\label{sec:Results}
By parallelizing the ArgMax calculation on the GPU, the runtime of this layer can be drastically reduced on the NVIDIA TX2 board compared to common CPU implementations. For an input image with a resolution of $\SI{640}{\px} \times \SI{400}{\px}$, the runtime can be reduced from $\SI{92}{\milli\second}$ to $\SI{0.05}{\milli\second}$. A comparison of the performance for the respective CPU and GPU implementation of the ArgMax layer can be found in Table~\ref{tab:ArgMaxComparison}. For larger image resolutions, the factor is even greater, since the parallelization can be better utilized.

The channel pruning method applied to the ENet allowed us to reduce the required GFLOPs from 1.87 to 1.34 for an input size of $\SI{640}{\px} \times \SI{400}{\px}$ as shown in Table~\ref{tab:Hardware_requirements}. Furthermore, we were able to reduce the number of parameters from 363\,k to 255\,k. Accordingly, the required memory of the parameters decreases from $\SI{1.49}{\mega\byte}$ to $\SI{1.06}{\mega\byte}$ (FP32). Due to the lower number of FLOPs, the inference time could be increased on the CPU by 17.7\%. After all, a runtime improvement from $\SI{11.06}{\fps}$ to $\SI{11.53}{\fps}$ could be achieved on the GPU. All computational measurements were done on the NVIDIA TX2 board with CUDA~9.0 and cuDNN~7.0.

To compare quality results we used mean intersection over union (mIoU) and global accuracy. The mIoU of our 20 classes dropped from 53.8\% to 51.4\% of our custom fisheye test dataset. The IoU values for each class are shown in Table~\ref{tab:ENet_IOU}. Interestingly, the IoU decreases especially for classes with a small pixel density (e.g.~pole). For classes with a high pixel density, the value remains the same or even increases (e.g. road), which explains the slightly increased global accuracy, which has increased from 94.04\% to 94.12\%.

{
\renewcommand{\arraystretch}{1.2}\setlength{\tabcolsep}{0.5em}
 \begin{table}[ht] 
 \caption{Comparison of performance using CPU and GPU implementation of ENet for an input image size of $\SI{640}{\px} \times \SI{400}{\px}$.}
 \centering
 \begin{tabular}{c|c|c}
    \textbf{Network} & \textbf{ArgMax layer} & \textbf{Performance} \\
  \hline
     CPU & CPU & $ \makebox[4em]{\hfill\SI{0.23}{\fps}}$ \\
     GPU & CPU & $ \makebox[4em]{\hfill\SI{5.46}{\fps}}$ \\
     GPU & GPU & $ \makebox[4em]{\hfill\SI{11.06}{\fps}}$ \\
 \end{tabular}
  \label{tab:ArgMaxComparison}
 \end{table}}
 
{
\renewcommand{\arraystretch}{1.2}\setlength{\tabcolsep}{0.3em}
 \begin{table}[ht] 
 \caption{Comparison of hardware requirements and performance of ENet (including ArgMax calculation) for an input image size of $\SI{640}{\px} \times \SI{400}{\px}$.}
 \centering
 \begin{tabular}{c|c|c|c|c|c}
    \thead{Modell\\~} & \thead{GFLOPs\\~} & \thead{Paramter\\~} & \thead{Model size \\ (FP32)} & \thead{Performance\\CPU} & \thead{Performance\\GPU}\\
  \hline
     ENet        & 1.87 & 363\,k & $\SI{1.49}{\mega\byte}$ & $\SI{0.23}{\fps}$ & $\makebox[4em]{\hfill\SI{11.06}{\fps}}$\\
     ENet pruned & 1.34 & 255\,k & $\SI{1.06}{\mega\byte}$ & $\SI{0.28}{\fps}$ & $\makebox[4em]{\hfill\SI{11.53}{\fps}}$\\
 \end{tabular}
  \label{tab:Hardware_requirements}
 \end{table}}

{
\renewcommand{\arraystretch}{1.2}\setlength{\tabcolsep}{0.5em}
 \begin{table}[ht]
 \caption{Representation of the Intersection over Union (IoU) values per class for comparison of ENet before and after pruning.}
 \centering
 \begin{tabular}{c|c|c}
    \textbf{Classes} & \textbf{IoU ENet} & \textbf{IoU ENet pruned} \\
  \hline
     Road & 95.4\% & 95.6\%\\
     Sidewalk & 75.1\%& 75.2\%\\
     Building & 87.3\%& 87.2\%\\
     Wall & 65.3\%& 63.8\%\\
     Fence & 45.1\%& 42.7\%\\
     Pole & 30.7\%& 23.4\%\\
     Traffic light & 41.5\%& 39.1\%\\
     Traffic sign & 27.1\%& 24.9\%\\
     Vegetation & 80.1\%& 80.2\%\\
     Terrain & 24.6\%& 25.0\%\\
     Sky & 95.9\%& 96.1\%\\
     Person & 42.2\%& 40.3\%\\
     Rider & 13.6\%& 07.3\%\\
     Car & 83.1\%& 82.2\%\\
     Truck & 40.3\%& 35.4\%\\
     Bus & 51.3\%& 47.3\%\\
     Motorcycle & 15.0\%& 07.5\%\\
     Bicycle & 47.9\%& 43.0\%\\
     Road markings & 61.2\%& 61.2\%\\
   \hline  
     \textbf{Mean IoU} & \textbf{53.8\%} & \textbf{51.4\%}\\
 \end{tabular}
  
  \label{tab:ENet_IOU}
 \end{table}}

\section{Conclusion}
We have proposed a parallelized ArgMax Layer implementation that dramatically improves runtime for semantic segmentation models. For an input image size of $\SI{640}{\px} \times \SI{400}{\px}$, the runtime of the ArgMax layer on the NVIDIA TX2 board could be reduced by a factor of $\num{1840}$.

In addition, the hardware requirements for the ENet could be significantly reduced by channel pruning. The number of required GFLOPs could be reduced by about 30\%, which allows a theoretical speed up of 1.4.

Despite this significant reduction in the hardware requirements of the already efficient ENet model, the IoU value has fallen slightly only for small classes (e.g.~pole).
These results are essential for embedded systems to use semantic segmentation in a real-time capable application such as the generation of the semantically segmented birds-eye view.
In the future, we plan to extend the channel pruning method to additional layers of the ENet to further reduce hardware requirements. Furthermore, we plan to work on a more comprehensive fine-tuning step to maintain the quality of the ENet at a similar level as before pruning.

\section*{Acknowledgments}

We would like to thank Senthil Yogamani and our colleagues at Valeo Vision Systems in Ireland for collaboration on our dataset using automotive fisheye cameras. We would like to thank Valeo, especially J\"org Schrepfer, for the opportunity doing fundamental research.

\newpage

%
%
\bibliographystyle{splncs}
\bibliography{ref.bib}


\end{document}